\documentclass[letterpaper]{article} % DO NOT CHANGE THIS
\usepackage{tikz}
\usepackage{aaai23}  % DO NOT CHANGE THIS
\usepackage{times}  % DO NOT CHANGE THIS
\usepackage{helvet}  % DO NOT CHANGE THIS
\usepackage{courier}  % DO NOT CHANGE THIS
\usepackage[hyphens]{url}  % DO NOT CHANGE THIS
\usepackage{graphicx} % DO NOT CHANGE THIS
\usepackage{natbib}  % DO NOT CHANGE THIS AND DO NOT ADD ANY OPTIONS TO IT
\usepackage{caption} % DO NOT CHANGE THIS AND DO NOT ADD ANY OPTIONS TO IT
\usepackage{algorithm}
\usepackage{algorithmic}

\usepackage{newfloat}
\usepackage{listings}
\DeclareCaptionStyle{ruled}{labelfont=normalfont,labelsep=colon,strut=off} % DO NOT CHANGE THIS
\floatstyle{ruled}
\newfloat{listing}{tb}{lst}{}
\floatname{listing}{Listing}
\title{LowDINO - A Low Parameter Self Supervised Learning Model}
\author {
    Sai Krishna Prathapaneni,\textsuperscript{\rm 1}
    Shvejan Shashank, \textsuperscript{\rm 2}
     Srikar Reddy K \textsuperscript{\rm 3}
}
\affiliations {
    \textsuperscript{\rm {1,2}} Department of ECE, New York University \\
    \textsuperscript{\rm {3}}Indian Insitute of Technology, Kharagpur\\
     sp7238@nyu.edu,  ssm10076@nyu.edu,srikark30@gmail.com \url{https://github.com/saikrishna-prathapaneni/LowDINO}
}

\usepackage{bibentry}
\begin{document}

\maketitle

\begin{abstract}

This research aims to explore the possibility of designing a neural network architecture that allows for small networks to adopt the properties of huge networks, which have shown success in self-supervised learning (SSL), for all the downstream tasks like image classification, segmentation, etc. Previous studies have shown that using convolutional neural networks (ConvNets) can provide inherent inductive bias, which is crucial for learning representations in deep learning models. To reduce the number of parameters, attention mechanisms are utilized through the usage of MobileViT blocks, resulting in a model with less than 5 million parameters. The model is trained using self-distillation with momentum encoder \cite{DBLP:journals/corr/abs-2006-07733} and a student-teacher architecture is also employed, where the teacher weights use vision transformers (ViTs) from \cite{oquab2023dinov2}. The model is trained on the ImageNet1k dataset. This research provides an approach for designing smaller, more efficient neural network architectures that can perform SSL tasks comparable to heavy models. 
\end{abstract}

\section{Introduction}

SSL has become popular in recent years, due to its ability to leverage large amounts of unlabeled data and improve the performance of models mainly in downstream tasks with its ability to learn efficient representations of the data. SSL, in the field of Natural language processing(NLP), performs well in the paradigms of contrastive and generative setups due its limited dimensionality \cite{BERT} \cite{GPT} and progressing into efficient models, on the other hand in computer vision (CV) a wide array of architectures have been proposed On extremely competitive benchmarks like as ImageNet, SSL approaches for computer vision have been able to match and in some cases even beat models trained on labeled dataset \cite{DBLP:journals/corr/abs-2201-05119} and have been excellent KNN classifiers \cite{DINO}. SSL applied to multi-modalities such as audio and videos \cite{Wickstr_m_2022} with Frames dropped in between to predict. The objective of SSL is to mask a certain part of an image or a word in the text and try to predict the masked region. 
The objective of predicting the region leverages the model to capture complex relationships across various parts of an image or text without the necessity of providing any labels. In the research, we provide our results based on training from the data of CIFAR100 and ImageNet1k datasets. We use the methodologies of DINO where self-distillation of the model is done with the help of a momentum-encoded teacher network, In addition, we also perform the Distillation LowDINO models to calculate the KNN accuracies and Linear accuracies.  

\section{Literature Survey}

One of the earliest SSL algorithms is  developed by \cite{DBLP:journals/corr/DoerschGE15} essentially predicts the relative position of two randomly chosen patches in an image. This concept has been overtaken by "jigsaw" algorithms \cite{DBLP:journals/corr/PathakKDDE16}, which divide an image into an array of discontinuous patches and estimate their relative positioning and one includes to predict the rotation angle of an Image\cite{RotNet}. Colorization-based SSL methods train to predict the original RGB values from grayscale \cite{DBLP:journals/corr/ZhangIE16}. \cite{DBLP:journals/corr/PathakGDDH16} trains a model to predict object motion in a single frame and then modify the resultant features to deal with single-frame detection challenges.  \cite{DBLP:journals/corr/AgrawalCM15} predicts the ego-motion of a camera
given multiple frames. \cite{DBLP:journals/corr/Owens0MFT16} proposes to remove the audio track from a video, and then predict the missing sound. In Multiview variance, One of the most widely used methods for learning from unlabeled data involves labeling images with pseudo labels using a weakly trained network, then training using these labels in a conventional supervised manner \cite{Lee2013PseudoLabelT}.

\begin{figure}[t]
\centering
\includegraphics[width=0.9\columnwidth]{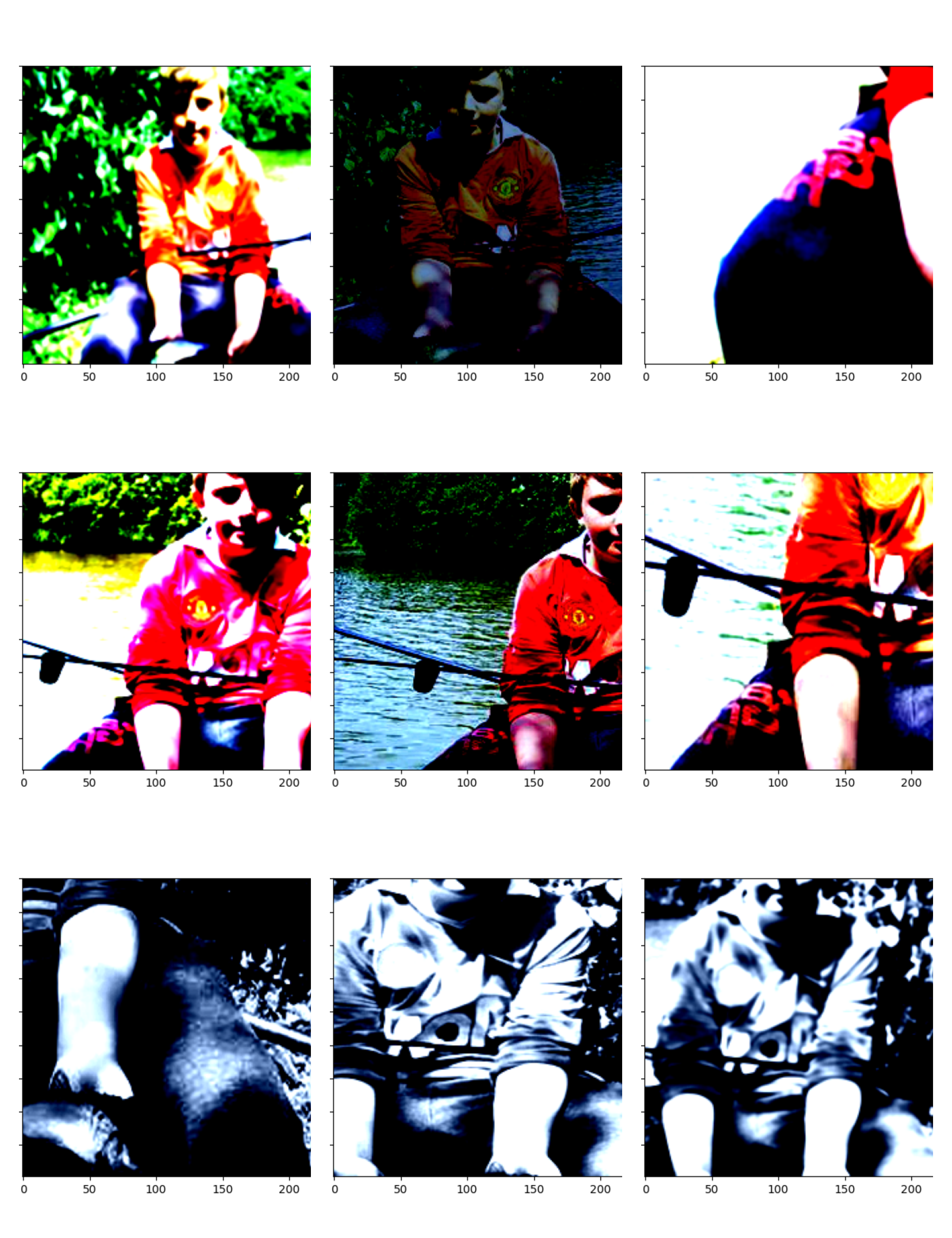} % Reduce the figure size so that it is slightly narrower than the column. Don't use precise values for figure width. This setup will avoid overfull( boxes.
\caption{ Crops generated per Image (resized to 224x224)}
\label{fig1}
\end{figure}

 Though these contrastive-based techniques have gained interest and generated better representations, non-contrastive-based approaches where student-teacher-based networks in recent years have been showing comparative results to that of supervised and semi-supervised alternative paradigms.
\begin{table}[t]
\centering
%\resizebox{.95\columnwidth}{!}{
\begin{tabular}{l|l|l}
    \textbf{Model} & \textbf{Parameter} & \textbf{KNN} \\
    \textbf{MobileVIT(LowDINO)}& 5.5M & 53.05\%  \\
    \textbf{ResNet50(DINO)} & 23.5M & 50.04\%  \\
    \textbf{ViT\_S(DINOv2)} & 21.5M & 55.64\%  \\
    \textbf{ResNet5M(LowDINO)} & 4.9M & 34.3\%  \\
    \textbf{MobileNet(ImageNet)} & 3.5M & 32.14\%  \\
    \textbf{ResNet50 (BYOL)} & 23.5M & 50.14\%  \\
   
\end{tabular}
\caption{KNN Accuracy on CIFAR10 with out finetuning}
\label{table1}
\end{table}

\begin{table}[t]
\centering
\begin{tabular}{l|l}
\textbf{Parameter} & \textbf{Value}\\
batch\_size & 64 \\
logging\_freq & 1 \\
n\_crops & 4 \\
n\_epochs & 100 \\
out\_dim & 1024 \\
optim & SGD \\
clip\_grad & 2.0 \\
norm\_last\_layer & False \\
batch\_size\_eval & 8 \\
teacher\_temp & 0.04 \\
student\_temp & 0.1 \\
device\_ids & [0] \\
pretrained & True \\
lr & 0.0005 \\
min\_lr & 1e-06 \\
warmup\_epochs & 10 \\
weight\_decay & 0.04 \\
weight\_decay\_end & 0.4 \\
momentum\_teacher & 0.9995 \\
\end{tabular}
\caption{Hyper parameters for the Experiment for LowDINO. Showing experiment results with MobileVit small with 5.5M parameters}
\label{table2}
\end{table}

\begin{table*}[t]
\centering

\begin{tabular}{l|l|l|l}
 \textbf{Model} &
\textbf{ KNN accuracy} &
 \textbf{Linear 10\% data} &
\textbf{Linear 30\% data} \\
MobileVit(LowDINO) &
53.05\% &
61.93 \% &
66\% \\
ResNet5M(LowDINO)&
34.3\% &
39.12\% &
41.21 \% \\
ResNet50(DINO)&
50.04\% &
78.48\% &
79.72 \% \\
Vit\_S14(DINOv2)&
55.04\% &
67.1\% &
69.9\% \\
Mobilenet(INet1k)&
32.14\% &
56.7\% &
59.9\%
\end{tabular}
%}
\caption{Model fine tuned for 30 epochs on CIFAR10 with 10\% and 30\% of data}
\label{table3}
\end{table*}

The recent advancement with DINOv2 \cite{oquab2023dinov2}, which is an extension from DINO \cite{DINO} shows a promising path in the direction of SSL, with Imagenet1k top1 accuracy reaching $\approx 89\%$ showing a path to adopt. DINO  extends from  the extension of two other self-supervised learning algorithms, BYOL (Bootstrap Your Own Latent)\cite{BYOL} and SimSiam (Simplified Self-Supervised Learning with Contrastive Predictive Coding) \cite{SIMSAM}, is designed to work with discrete representations and  relies on a momentum encoder and centering to avoid modal collapse. The relevance of the research paper lies in its exploration of the limitations of high-accuracy models based on supervised learning for image classification and segmentation applications, which require large amounts of labeled training data. The paper addresses the challenge of working with limited labeled data in certain applications, such as medical and astronomical imaging, by proposing the use of self-supervised learning (SSL). The paper highlights the benefits of SSL and provides evidence of the effectiveness of the DINOv2 model for image classification experiments on the ImageNet1k \cite{ILSVRC15} and extends itself from adopting properties from \cite{IBOT}. However, it also acknowledges the challenges posed by highly parameterized models like DINOv2, which have large computational and memory requirements and are unsuitable for low-powered devices. As such, the paper's relevance extends to the development of efficient models that can perform SSL tasks with lower computational requirements, making them more suitable for low-powered devices like embedded systems, mobile devices, and IoT devices. By addressing these challenges, the research paper contributes to the advancement of SSL in image classification, segmentation, and related applications, with the potential to impact various industries, including healthcare, robotics, and autonomous vehicles. One such attempt \cite{gao2022disco} Distilled Contrastive Learning (DisCo) by constraining student model to teacher embedding dimension.
Our research work aims to explore ways to reduce the model’s complexity without a significant decrease in performance as compared to recent SSL implementations. We trained a low parameter model with a similar training paradigm as DINO with MobileVIT blocks\cite{mobilevit} resulting in 4.9 million parameter backbone and distilling it into a series of smaller models such as MobileNets\cite{Mobilenet} and ResNets\cite{resnet} \cite{Residual_network}.
\begin{figure}[t]
\centering
\includegraphics[width=0.9\columnwidth]{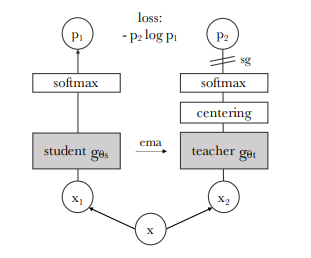} % Reduce the figure size so that it is slightly narrower than the column. Don't use precise values for figure width. This setup will avoid overfull( boxes.
\caption{ Architecture in DINO\cite{DINO}}
\label{fig2}
\end{figure}

\begin{figure}[t]
\centering
\includegraphics[width=0.9\columnwidth]{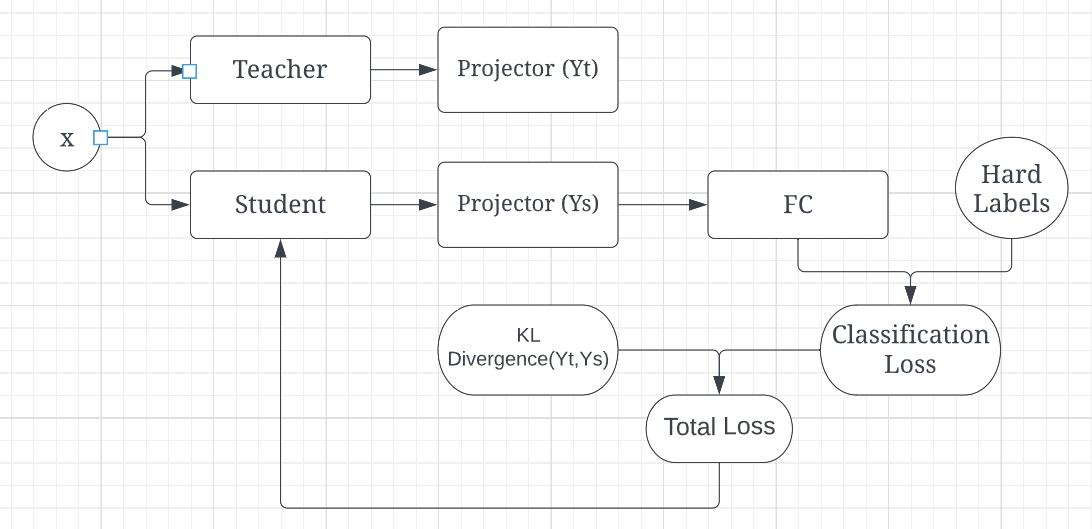} % Reduce the figure size so that it is slightly narrower than the column. Don't use precise values for figure width.This setup will avoid overfull( boxes.
\caption{ Architecture used to train via offline distillation of LowDINO models}
\label{fig3}
\end{figure}

\section{Methodology}
The project aims to reduce the model size to less than 5 million parameters by considering the low-parameter architecture and reducing the size of DINO, and DINOv2 models considerably low.
The architecture employed as shown in Figure \ref{fig2} is with a replaceable backbone network, which was tested with MobileVit small networks and ResNet with 5M parameters approximately. 
We made use of properties from \cite{DINO} and \cite{oquab2023dinov2} to construct our model. Table \ref{table2} shows the hyperparameters selected for the training of MobileViTs where the logging frequency is set to 1, indicating that the progress is logged after every training iteration. cilp\_grad is the maximum value allowed for gradient elements during training to prevent gradient explosion. norm\_last\_layer is a flag indicating whether to apply normalization to the output layer of the head. In this case, the last layer normalization is set to False. batch\_size\_eval is the number of evaluation examples processed in one forward pass during each evaluation iteration. Here, the batch size for evaluation is set to 8. teacher\_temp is used in the distillation process that controls the softness of the teacher's predictions. student\_temp is similar to the teacher temperature but used for the student model. Here, a temperature of 0.1 is used. warmup\_epoch is the number of initial epochs where the learning rate is gradually increased from the minimum learning rate to the initial learning rate. weight\_decay\_end is the weight decay value at the end of the training. Here, weight decay ends at 0.4. Teacher momentum is the momentum value used in the momentum encoder of the self-distillation process. Similar \cite{DINO} we generated 4 and 10 crops for MobileVit and ResNet5m\cite{resnet5m} respectively of the same image i.e $  x_1, x_2 \in crops(x) $ out of the $n$ local crops generated $2$ global crops consisting of scale between $(0.4, 1)$ and $n-2$ local crops with a scale between $(0.05, 0.4)$.where crops are generated based on the following augmentations. 

\begin{itemize}
    \item \textbf{Gaussian Blur }:50\% of the time with a radius of 0.1 to 0.5 for local crops and 10\% of the time in global crops.
    \item \textbf{Solarization}:20\% of the times in global transforms. 
    \item \textbf{RandomCrop}:$global crops \in (0.4,1)$ and $local crops \in (0.05,0.4)$ 
    \item \textbf{Resize}:Bicubic interpolation to actual size of 224 x224 for local crops
    \item \textbf{Color jitter}: Applied for 80\% of the time, changed parameters of color, contrast, brightness Grayscale(p=0.2)and saturation.
    \item \textbf{Flip}: Image flip horizontal and vertical flip of the images
\end{itemize}

After the generation of crops global crops are sent to the teacher model and local and global crops are sent to the student network. 

\begin{equation}\label{weightupdate}
\theta_{t+1} \leftarrow m \theta_{t} + (1-m) \theta_{s},
\end{equation}

\begin{equation}\label{cosineschedule}
\eta_t = \eta_{\mathrm{min}} + \frac{1}{2} (\eta_{\mathrm{max}} - \eta_{\mathrm{min}})(1 + \cos(\frac{T_c - t}{T_c}\pi))
\end{equation}

 Weights are updated based on the teacher based on equation \ref{weightupdate} where $\theta_t$ and $\theta_s$ are the parameters of the target (teacher) and online (student) networks respectively, and $m$ is the momentum factor. This operation updates the target network parameters by taking a weighted average between the current target network parameters and the current online network parameters. The momentum parameter(m) is scheduled based on the cosine annealing schedule shown in equation \ref{cosineschedule} where $\eta_{\mathrm{min}}$ is the minimum learning rate, $\eta_{\mathrm{max}}$ is the maximum learning rate, $t$ is the current training iteration, $T_c$ is the total number of training iterations, and $\cos(\frac{T_c - t}{T_c}\pi)$ is a cosine function that varies smoothly between -1 and 1 over the course of the training schedule. To avoid the modal collapse between the teacher and student we introduce a centering and momentum encoder to update the teacher model from the student based on equation \ref{weightupdate}. Finally, the loss is a calculation between the probability distributions. unlike in DINO/DINOv2, we limited our dimensionality to 1024 to compute the softmax outputs.

In Table: \ref{table1} we show the KNN accuracy\cite{KNN} of two different backbone networks ResNet with 4.9M parameters, and MobileVit with 5.5M parameters. We obtain comparable results for training the model from ImagNet1k constrained to 150K images of data, for 100 epochs, as shown in table \ref{table1} for our top performing model on the CIFAR10 dataset we obtain an accuracy of 53\% outperforming DINO's ResNet50 model with 4times low parameters.

\begin{equation} \label{eq:softmax}
P_{s}(x)_{i} = \frac{\exp\left(\frac{g{\theta_s}(x)i}{\tau_s}\right)}{\sum_{k=1}^{K} \exp\left(\frac{g_{\theta_s}(x)_k}{\tau_s}\right)}
\end{equation}

where $P_s(x)_i$ represents the probability of the $i^{th}$ class for sample $x$ under the softmax function $s$, $g_{\theta_s}(x)_i$ is the $i^{th}$ element of the function $g_{\theta_s}(x)$, $\tau_s$ is the temperature parameter, and $K$ is the output dimension. Loss is computed between the logits of the model of student and teacher models and updating just the student models at the end. At the end of the training we use backbone model for the downstream tasks.

\subsection{Distillation}
Model distillation is a popular technique for transferring knowledge from a large, complex model to a smaller, more efficient one. The process involves training a smaller model, called a student model, to replicate the behavior of a larger, more complex model, called a teacher model. The student model is trained on the same task and dataset as the teacher model, but with the added guidance of the teacher's predictions. The technique was first introduced in \cite{hinton2015distilling}. in the paper "Distilling the Knowledge in a Neural Network" and has since been widely adopted in the deep learning community.
    In a similar paradigm we introduce offline distillation \cite{gou2021knowledge} and SimKD \cite{chen2022knowledge} to distill into students where teacher weights are not updated but student weights are based on the KL divergence between logits of teacher and student networks are used. In addition, we introduce classification loss into the system with hard labels similar to equation \ref{vanilla distillation} following show the networks used as a part of the distillation. Detailed architecture implemented for the distillation process is portrayed in the figure: \ref{fig3}. Figure \ref{fig7} shows the accuracies generated of KNN after training with the architecture mentioned
\begin{equation}\label{vanilla distillation}
L = (1-\alpha)CE(y, \hat{y}) + \alpha KL(T(y), T(\hat{y}))
\end{equation}

\begin{figure}[t]
\centering
\includegraphics[width=0.9\columnwidth]{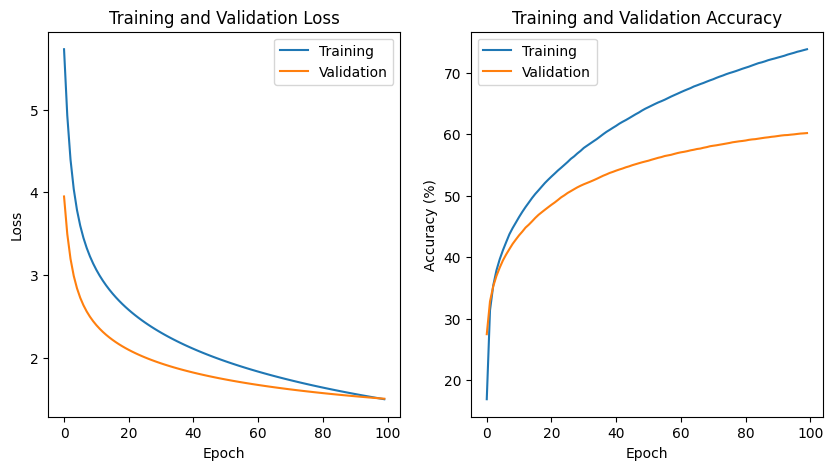} % Reduce the figure size so that it is slightly narrower than the column. Don't use precise values for figure width. This setup will avoid overfull( boxes.
\caption{ ResNet50 backbone(DINO) fine-tune on CIFAR 100, backbone frozen, FC updated}
\label{fig4}
\end{figure}

\begin{figure}[t]
\centering
\includegraphics[width=0.9\columnwidth]{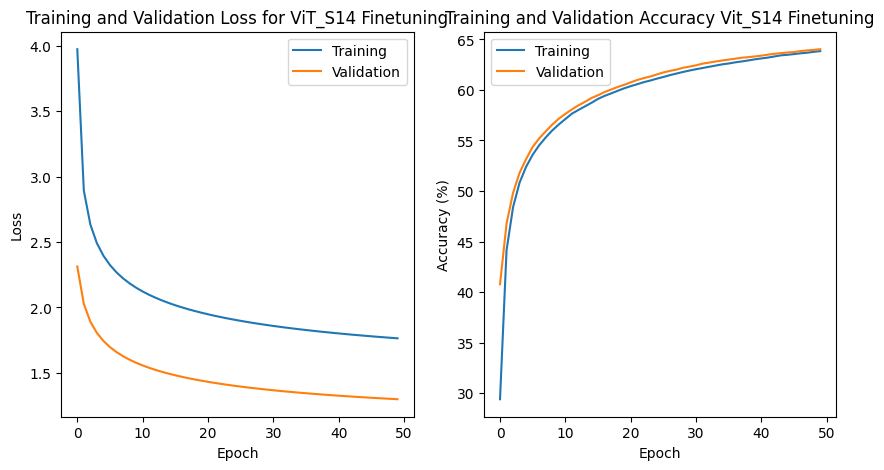} % Reduce the figure size so that it is slightly narrower than the column. Don't use precise values for figure width. This setup will avoid overfull( boxes.
\caption{ ViT\_S14(DINOv2) fine-tune on CIFAR 100, backbone froze, FC updated }
\label{fig5}
\end{figure}

where $CE(y, \hat{y})$ is the cross-entropy loss between the ground truth labels $y$ and the predicted labels $\hat{y}$, $KL(T(y), T(\hat{y}))$ is the KL divergence between the softened probabilities $T(y)$ generated by the teacher model and the softened probabilities $T(\hat{y})$ predicted by the student model, and $\alpha$ is a hyper parameter that controls the balance between the two loss terms. The process implemented, Though the process involves the usage of cross entropy loss from the labels, we removed the hard loss from the procedure and just used soft labels to compute the loss and update the weights of students.
following are the procedures used to train the model. we post a relative increase of $ \approx 10\%$ for ResNet5m and approximately equal amount for MobileViT from the baseline training.
\begin{itemize}
    \item DINO ResNet50 distills ResNet5M
    \item DINOv2 Vit\_S distills MobileVit\_S
\end{itemize}

\section{Results and conclusion}
    The research aimed to show low parameter models can also learn better representations when trained with low dimensionality of only 1024 considered here and without a big drop in accuracy. The results compiled for ResNet5M and MobilVit\_S model with self-distillation and offline distillation, from table \ref{table3}, we can observe that although the best performing model is still DINOv2, with 21.5M parameters, the MobileVit(LowDINO) with 5.5M parameters is very close in terms of KNN evaluation, A similar trend can be observed after fine-tuning over 10 \% and 30\% data of CIFAR10, MobileVit(LowDINO) performs very close to DINOv2 while  having almost 4 times less parameters and outperforming ResNet50 model of DINOv1 model. While these results seem promising in terms of parameter reduction, An extensive exploration needs to be done.

\section{Future Scope}
While the results posted here are promising with the amount of data used which is restricted to ImageNet1k with dataset sizes of 150k, 300k, and CIFAR100 datasets. It is worth exploring more further down the lane with similar paradigms of training by applying other loss functions and using larger dataset sizes including Imagenet22k/ LVM142M datasets and aggresive abblation studies are required to conclude on the final models best performing models.

\begin{figure}[t]
\centering
\includegraphics[width=0.9\columnwidth]{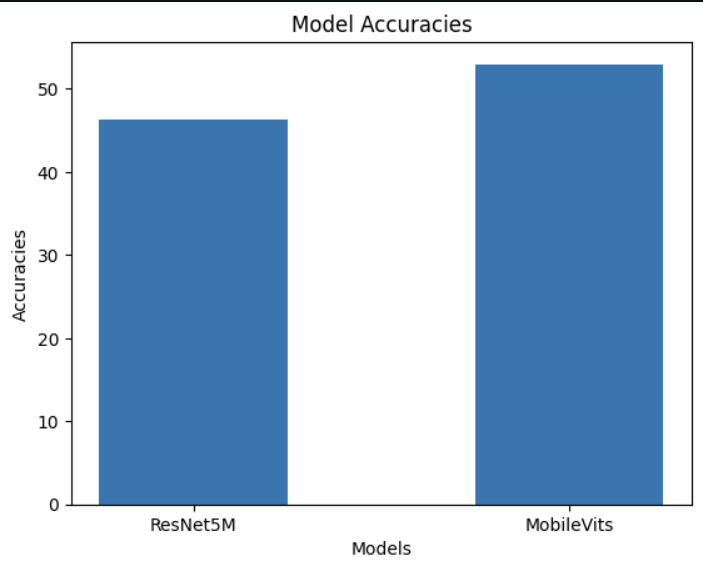} % Reduce the figure size so that it is slightly narrower than the column. Don't use precise values for figure width. This setup will avoid overfull( boxes.
\caption{ KNN accuracies generated for the models using distillation, Models use CIFAR100 Dataset to train}
\label{fig7}
\end{figure}

\bibliography{aaai23}

\end{document}